\documentclass[runningheads]{llncs}
\usepackage[T1]{fontenc}
\usepackage{graphicx,verbatim}
\usepackage{booktabs}
\usepackage{multirow}
\usepackage{amsmath,amssymb,amsfonts}
\usepackage{bm}
\usepackage{mathtools}
\usepackage{array}
\usepackage{tabularx}
\usepackage{url}
\usepackage{hyperref}
\hypersetup{
    colorlinks=true,
    linkcolor=blue,
    citecolor=blue,
    urlcolor=blue
}
\usepackage{enumitem}
\usepackage[misc]{ifsym}
\usepackage[table]{xcolor}

\begin{document}

\title{Task-Adaptive 3D Cross-Field MRI Translation via Field-Conditioned Content-Style Pretraining}
\titlerunning{Task-Adaptive 3D Cross-Field MRI Translation}

\author{
Haowen Pang\inst{1}
\and
Yingqi Hao\inst{2}
\and
Pengli Zhu \inst{3}$^{(\textrm{\Letter})}$
}
\authorrunning{H. Pang et al.}
%
\institute{
School of Integrated Circuits and Electronics, Beijing Institute of Technology, Beijing, China \\
\and
School of Biomedical Engineering, Tsinghua Medicine, Tsinghua University, Beijing, China  \\
\and
Department of Electronic Engineering, The Chinese University of Hong Kong, Hong Kong, China \\
\email{penglizhu@cuhk.edu.hk}\\
}


  
\maketitle              

\begin{abstract}
Magnetic field strength is a major source of domain shift in magnetic resonance imaging (MRI), affecting signal-to-noise ratio, tissue contrast, spatial detail, and the visibility of anatomical boundaries. The MRIxFields 2026 challenge investigates this problem through cross-field MRI translation across acquisitions at 0.1T, 1.5T, 3T, 5T, and 7T. Its three tasks, Any-to-7T, 0.1T-to-High, and Any-to-Any synthesis, require the generation of target-field image characteristics while preserving subject-specific anatomy. This problem is particularly challenging because paired acquisitions of the same subject across multiple field strengths are rarely available for training.
We propose a 3D unpaired cross-field MRI translation framework based on field-conditioned content-style pretraining. The proposed framework first learns controllable field-to-field translation across all available field strengths by disentangling anatomical content from field-dependent contrast characteristics. The pretrained backbone is then adapted to task-specific target domains. Our model comprises a 3D content encoder, a 3D style encoder, a field-conditioned style generator, an AdaIN-modulated decoder, and a multi-field discriminator. Adversarial learning encourages realistic target-field appearance, while cycle-consistency, identity, content, style, and diversity constraints promote anatomical fidelity and controllable translation.
We evaluate the proposed method on MRIxFields data spanning five field strengths and three MRI modalities. Experiments on paired test data demonstrate that the framework can adapt to the three challenge settings while preserving three-dimensional anatomical structure in the synthesized volumes.
The implementation code are publicly available at \href{https://github.com/Idea89560041/3D-MRI-Field-Translation}{https://github.com/Idea89560041/3D-MRI-Field-Translation}.

\keywords{Field strength \and MRI harmonization \and Unsupervised learning}

\end{abstract}

\section{Introduction}

Magnetic field strength introduces a fundamental trade-off between MRI accessibility and image quality~\cite{marques2019lowfield,sarracanie2015lowcost}. Ultra-high-field MRI, particularly 7T imaging, provides improved signal-to-noise ratio and contrast-to-noise ratio, enabling high-resolution anatomical imaging and enhanced visualization of subtle brain structures~\cite{kraff2015mri,ladd2018pros}. However, 7T systems remain expensive, technically demanding, and substantially less accessible than conventional clinical scanners. In contrast, low-field and ultra-low-field MRI systems can reduce acquisition cost and improve portability, but their lower field strengths generally limit signal-to-noise ratio, spatial resolution, and diagnostic detail~\cite{marques2019lowfield,sarracanie2015lowcost}. Cross-field MRI translation therefore has the potential to facilitate low-field image enhancement, Any-to-7T synthesis, and retrospective harmonization of heterogeneous MRI archives.

The MRIxFields 2026 challenge formalizes this problem across five field strengths and three brain MRI modalities~\cite{MRIxFields2026}. Its evaluation criteria jointly emphasize target-field realism, anatomical fidelity, quantitative consistency, perceptual quality, and regional structural preservation. These requirements make cross-field MRI synthesis more challenging than conventional image style transfer. An effective model must modify field-dependent image appearance while preserving subject-specific anatomical structures throughout the three-dimensional volume.

Deep learning has substantially advanced medical image translation in recent years~\cite{pang2026generative}.
Early approaches commonly formulated translation as supervised image-to-image regression, using convolutional neural networks to directly map source images to target images through voxel-wise reconstruction objectives~\cite{pang2026cascaded}. Generative adversarial networks (GANs) subsequently improved perceptual realism by combining reconstruction objectives with adversarial learning~\cite{pang2023ncct,pang2026drcam,zhang2025breathvisionnet,zhu2025q}.
More recently, diffusion-based generative models have achieved strong performance in medical image synthesis by progressively learning the target image distribution and enabling high-fidelity generation under image-conditional guidance~\cite{pang2025d,pang2025cascaded,zhu2025cycle}.
Despite these advances, most existing methods are developed for a fixed source-target mapping, require paired supervision, operate slice by slice, or focus on modality translation.

Cross-field MRI translation remains challenging for several reasons. First, magnetic field strength affects not only global intensity distributions but also local tissue contrast, noise characteristics, spatial resolution, and the visibility of fine anatomical structures~\cite{shinohara2014,guan2022fast}. Second, paired scans acquired from the same subject across all field strengths are rarely available, which limits the feasibility of fully supervised high-field synthesis in realistic multi-field settings~\cite{bahrami2016reconstruction}. Third, many existing translation methods operate on two-dimensional slices or primarily focus on scanner, site, or vendor harmonization~\cite{cackowski2023imunity,liu2024learning,wu2025unpaired,wu2026unpaired}. Recent advances, such as invertible flow-based MRI harmonization, have improved multi-site and multi-modality alignment~\cite{zhu2026ihf}. Nevertheless, challenge-oriented cross-field synthesis additionally requires full-volume spatial consistency, controllable target-field appearance, and robust adaptation across multiple translation endpoints.

To address these challenges, we formulate the MRIxFields tasks as related instances of a unified 3D multi-domain translation problem. The three challenge settings, Any-to-7T, 0.1T-to-High, and Any-to-Any, share the same fundamental objective: preserving subject-specific anatomical content while transforming field-dependent image contrast and quality. This formulation motivates a representation-learning framework based on content-style disentanglement. Prior content-style representation learning methods have shown promise in preserving domain-invariant structural information while modifying domain-specific degradations in unpaired image enhancement settings~\cite{zhu2023unsupervised1,zhu2023unsupervised2}. Building on this idea, we first learn a unified 3D field-to-field translation backbone across all available field strengths, rather than independently training separate translators for each source-target pair. The pretrained backbone is subsequently adapted to task-specific target distributions.

In summary, this work formulates cross-field MRI synthesis as a unified 3D conditional multi-domain translation problem spanning 0.1T, 1.5T, 3T, 5T, and 7T acquisitions. We introduce a field-conditioned content-style pretraining strategy to disentangle subject-specific anatomical content from field-dependent image appearance, and adapt the resulting field-to-field translation backbone to the Any-to-7T, 0.1T-to-High, and Any-to-Any tasks. The proposed framework generates full-volume MRI outputs and is comprehensively evaluated using paired analyses from task-wise, modality-wise, and translation-direction-wise perspectives.

\section{Methodology}
\label{sec:method}

\subsection{Problem Formulation and Framework Overview}

Let $\mathcal{B}=\{0.1\mathrm{T},1.5\mathrm{T},3\mathrm{T},5\mathrm{T},7\mathrm{T}\}$ denote the set of magnetic field-strength domains in MRIxFields.
Given a source MRI volume $\mathbf{x}_a$ acquired at field strength $a\in\mathcal{B}$, the goal is to synthesize a target-field volume $\hat{\mathbf{x}}_{b}$ for a target field strength $b\in\mathcal{B}$.
The synthesized image should preserve the anatomical content of $\mathbf{x}_a$ while matching the field-dependent appearance of domain $b$.
During training, cross-field images are unpaired.
Thus, a sampled source image $\mathbf{x}_a$ and a sampled target-domain image $\mathbf{x}_b$ generally do not correspond to the same subject.
Paired cross-field data are used only for testing and metric computation, and are not used for model optimization.

\begin{figure}[!t]
    \centering
    \includegraphics[width=\textwidth]{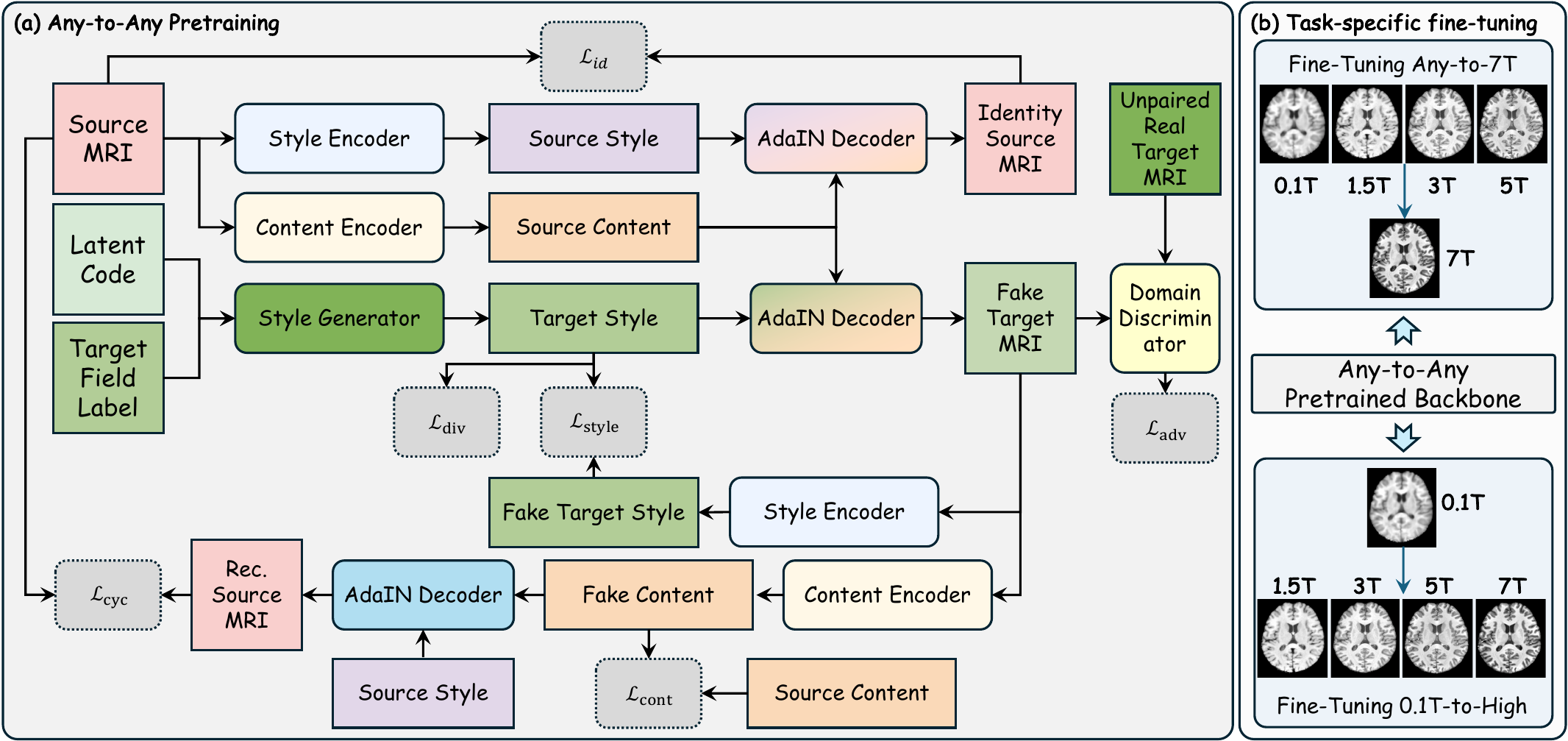}
    \caption{Overview of the task-adaptive 3D MRI field-strength translation framework. (a) Any-to-Any pretraining. Model optimization uses unpaired training patches; paired prospective data are used only for validation/testing and metric computation. (b) The pretrained Any-to-Any backbone supports Task 3 directly and is fine-tuned for Task 1 Any-to-7T synthesis and Task 2 0.1T-to-High synthesis.}
    \label{fig:framework}
\end{figure}

Fig.~\ref{fig:framework} illustrates the proposed 3D multi-domain translation framework.
We first train a generic any-to-any model across all field-strength domains.
The pretrained model is then adapted to task-specific source-target distributions for the challenge settings.
All models are trained on 3D patches and applied to full MRI volumes at inference using sliding-window translation.
Each MRI contrast modality, including T1W, T2W, and FLAIR, is trained independently.

\subsection{Any-to-Any Pretraining}

The proposed model disentangles each input MRI patch into a field-invariant content representation and a field-dependent style representation.
Given a source patch $\mathbf{x}_a$ with field label $a$, the content encoder $E_\mathrm{c}$ extracts an anatomical feature map $\mathbf{c}_a = E_\mathrm{c}(\mathbf{x}_a)$.
The content encoder is implemented as a 3D convolutional network with strided downsampling layers and residual blocks.
This design encourages $\mathbf{c}_a$ to preserve local volumetric anatomy while reducing field-specific intensity information.

A style encoder $E_\mathrm{s}$ extracts field-dependent contrast representations from real MRI patches.
It consists of a shared 3D convolutional trunk followed by field-specific linear heads.
Given an image patch $\mathbf{x}_a$ and its field label $a$, the corresponding style code is $\mathbf{s}_a = E_\mathrm{s}(\mathbf{x}_a,a)$.
The domain-specific heads allow the model to represent contrast characteristics associated with each magnetic field strength.

For target-conditioned synthesis, the model does not require a paired target image.
Instead, a mapping network $M$ maps a random latent vector $\mathbf{z}_b$ and a target field label $b$ to a target style code $\mathbf{s}_b = M(\mathbf{z}_b,b)$.
The generator $G$ combines the source content $\mathbf{c}_a$ with the target style $\mathbf{s}_b$ to synthesize a target-field patch $\hat{\mathbf{x}}_{a\rightarrow b}=G(\mathbf{c}_a,\mathbf{s}_b)$.
The generator uses 3D residual blocks modulated by adaptive instance normalization (AdaIN) \cite{huang2017arbitrary}, followed by trilinear upsampling and convolutional reconstruction layers.
A hyperbolic tangent activation constrains the output to the normalized intensity range.

In the any-to-any pretraining stage, source patches are sampled from all field-strength domains in $\mathcal{B}$.
For each source field $a$, a target field $b$ is sampled from $\mathcal{B}$, with non-identity translations used in most iterations.
A small probability of same-domain sampling is retained to regularize identity-preserving behavior.
The source patch $\mathbf{x}_a$ and target-domain patch $\mathbf{x}_b$ are sampled independently, so no paired supervision is required.

A field-conditioned discriminator $D_\phi$ encourages generated patches to match the target-domain distribution.
The discriminator is conditioned on the target field by concatenating a one-hot field map with the input patch.
The hinge discriminator loss is
\begin{equation}
    \mathcal{L}_{D} =
    \mathbb{E}_{\mathbf{x}_b}\left[\max(0,1-D_\phi(\mathbf{x}_b,b))\right]
    +
    \mathbb{E}_{\mathbf{x}_a,\mathbf{z}_b}\left[\max(0,1+D_\phi(\hat{\mathbf{x}}_{a\rightarrow b},b))\right].
\end{equation}
The adversarial loss for the generator is
\begin{equation}
    \mathcal{L}_{\text{adv}} =
    -\mathbb{E}_{\mathbf{x}_a,\mathbf{z}_b}
    \left[D_\phi(\hat{\mathbf{x}}_{a\rightarrow b},b)\right].
\end{equation}
Because adversarial distribution matching alone may alter anatomical structure, we impose cycle reconstruction.
The generated patch is re-encoded as
\begin{equation}
    \mathbf{c}_{a\rightarrow b}=E_\mathrm{c}(\hat{\mathbf{x}}_{a\rightarrow b}),
\end{equation}
and the source style $\mathbf{s}_a=E_\mathrm{s}(\mathbf{x}_a,a)$ is used to reconstruct the source patch:
\begin{equation}
    \tilde{\mathbf{x}}_a = G(\mathbf{c}_{a\rightarrow b},\mathbf{s}_a).
\end{equation}
The cycle loss combines voxel-wise and spatial-gradient consistency,
\begin{equation}
    \mathcal{L}_{\text{cyc}}
    =
    \|\tilde{\mathbf{x}}_a-\mathbf{x}_a\|_1
    +
    \lambda_{\nabla}
    \frac{1}{3}
    \sum_{d\in\{D,H,W\}}
    \left\|
    \nabla_d \tilde{\mathbf{x}}_a-\nabla_d \mathbf{x}_a
    \right\|_1 .
\end{equation}
We further use identity, content-consistency, and style-reconstruction losses.
The identity reconstruction is defined as
\begin{equation}
    \mathbf{x}^{\text{id}}_a =
    G(E_\mathrm{c}(\mathbf{x}_a),E_\mathrm{s}(\mathbf{x}_a,a)),
    \quad
    \mathcal{L}_{\text{id}}=\|\mathbf{x}^{\text{id}}_a-\mathbf{x}_a\|_1 .
\end{equation}
Content consistency encourages the translated patch to preserve the source anatomical representation:
\begin{equation}
    \mathcal{L}_{\text{cont}}
    =
    \left\|
    E_\mathrm{c}(\hat{\mathbf{x}}_{a\rightarrow b})
    -
    \mathrm{sg}(E_\mathrm{c}(\mathbf{x}_a))
    \right\|_1 ,
\end{equation}
where $\mathrm{sg}(\cdot)$ denotes stop-gradient.
Style reconstruction encourages the generated patch to express the sampled target style:
\begin{equation}
    \mathcal{L}_{\text{style}}
    =
    \left\|
    E_\mathrm{s}(\hat{\mathbf{x}}_{a\rightarrow b},b)
    -
    \mathrm{sg}(\mathbf{s}_b)
    \right\|_1 .
\end{equation}
To reduce style collapse, two latent vectors $\mathbf{z}_b$ and $\mathbf{z}'_b$ are sampled for the same target field.
The diversity objective is
\begin{equation}
    \mathcal{L}_{\text{div}}
    =
    -
    \left\|
    G(\mathbf{c}_a,M(\mathbf{z}_b,b))
    -
    G(\mathbf{c}_a,M(\mathbf{z}'_b,b))
    \right\|_1 .
\end{equation}
The full generator objective is
\begin{equation}
    \mathcal{L}_{G} =
    \lambda_{\text{adv}}\mathcal{L}_{\text{adv}}
    + \lambda_{\text{cyc}}\mathcal{L}_{\text{cyc}}
    + \lambda_{\text{id}}\mathcal{L}_{\text{id}}
    + \lambda_{\text{cont}}\mathcal{L}_{\text{cont}}
    + \lambda_{\text{style}}\mathcal{L}_{\text{style}}
    + \lambda_{\text{div}}\mathcal{L}_{\text{div}}.
\end{equation}

\subsection{Task-Specific Fine-Tuning}

After any-to-any pretraining, the same architecture and loss functions are reused for task-specific adaptation. 
For Task 1, the model is initialized from the pretrained checkpoint and fine-tuned with the target field fixed to $7\mathrm{T}$, using source patches from $\{0.1\mathrm{T},1.5\mathrm{T},3\mathrm{T},5\mathrm{T}\}$ and target patches from the $7\mathrm{T}$ domain. 
For Task 2, the source field is fixed to $0.1\mathrm{T}$ and target patches are sampled from $\{1.5\mathrm{T},3\mathrm{T},5\mathrm{T},7\mathrm{T}\}$, yielding a controllable family of higher-field enhancement mappings. 
Task 3 uses the any-to-any pretrained model directly without additional fine-tuning; source and target fields are selected from $\mathcal{B}$ at inference, with identity mappings omitted unless explicitly requested.

\subsection{Implementation Details}

The proposed framework was implemented in PyTorch~\cite{paszke2019pytorch}. 
All models were optimized using Adam~\cite{kingma2014adam} with $\beta_1=0$ and $\beta_2=0.99$. During Any-to-Any pretraining, the learning rate was set to $1\times10^{-4}$, and the model was trained for 50,000 optimization steps. The pretrained model was subsequently adapted to each task-specific setting using a learning rate of $5\times10^{-5}$ for 5,000 optimization steps. Mixed-precision training and distributed data parallelism were employed on NVIDIA RTX A6000 GPUs.

For Any-to-Any pretraining, the loss weights were set to $\lambda_{adv}=1$, $\lambda_{cyc}=10$, $\lambda_{id}=10$, $\lambda_{cont}=10$, $\lambda_{style}=10$, and $\lambda_{div}=0.1$. The gradient-consistency term within the cycle-consistency loss was weighted by $\lambda_{\nabla}=0.1$. During task-specific fine-tuning, the same loss formulation was retained, while $\lambda_{cyc}$, $\lambda_{id}$, and $\lambda_{div}$ were adjusted to 12, 8, and 0.05, respectively. Model checkpoints and qualitative preview volumes were saved every 500 optimization steps.

During training, the models were optimized on $64^3$ 3D patches sampled with a foreground-biased cropping strategy to increase the probability of observing anatomically informative regions.
At inference, full-resolution MRI volumes were translated using tiled 3D sliding-window inference. When overlapping patches were used, the predictions in overlapping regions were averaged to reduce boundary artifacts. For the reported test-time experiments, we used $128^3$ inference patches without overlap to improve computational efficiency.

The content encoder consisted of two downsampling stages followed by four residual blocks. The base channel width was set to 16, while the style-code dimension and latent-code dimension were set to 64 and 16, respectively. The generator employed AdaIN-modulated 3D residual blocks to inject field-dependent appearance information and used trilinear interpolation for upsampling. The discriminator was implemented as a field-conditioned 3D convolutional network, in which a one-hot field-strength map was concatenated with the input image patch.

During inference, the same target-field conditioning mechanism was used for all translation directions, allowing the model to change the requested output field without modifying the network parameters. 
The original image affine and header information were preserved when saving translated NIfTI volumes, ensuring spatial consistency with the source scans.

\section{Results}
\label{sec:results}

Following the challenge setting, training and validation data are treated as unpaired across field strengths.
Paired cross-field volumes are reserved for quantitative evaluation. All quantitative evaluations are performed on paired target-field references.
Let $\hat{x}$ be the generated image and $x$ be the paired target-field reference after robust normalization of the evaluated slice slab to $[-1,1]$.
We report normalized root mean squared error (nRMSE), mean absolute error (MAE), peak signal-to-noise ratio (PSNR), and structural similarity (SSIM) \cite{parida2024quantitative,wang2003multiscale}.

\subsection{Performance}
Table~\ref{tab:allresults} summarizes paired testing results by MRIxFields objective and modality.
Task 1 averages all non-7T source fields for Any-to-7T.
Task 2 averages all higher-field targets for 0.1T-to-High.
Task 3 averages all 20 non-identity directions for Any-to-Any.
Reporting modality-wise and task-level means separates overall performance from modality-specific failure modes.

\begin{table}[!t]
    \centering
    \caption{Paired testing results by MRIxFields synthesis objective and modality.}
    \label{tab:allresults}
    \begin{tabular}{lccccc}
        \toprule
        Task & Modality & nRMSE $\downarrow$ & MAE $\downarrow$ & PSNR $\uparrow$ & SSIM $\uparrow$ \\
        \midrule
        \multirow{3}{*}{Task 1: Any-to-7T}
        & T1W & 0.1076 & 0.1261 & 19.65 & 0.7278 \\
        & T2W & 0.0804 & 0.0834 & 21.96 & 0.7363 \\
        & FLAIR & 0.1007 & 0.1429 & 20.00 & 0.6359 \\

        \midrule

        \multirow{3}{*}{Task 2: 0.1T-to-High}
        & T1W & 0.0915 & 0.1227 & 21.26 & 0.7376 \\
        & T2W & 0.1012 & 0.0962 & 19.92 & 0.7076 \\
        & FLAIR & 0.1026 & 0.1188 & 20.09 & 0.6587 \\

        \midrule

        \multirow{3}{*}{Task 3: Any-to-Any}
        & T1W & 0.0881 & 0.0994 & 21.61 & 0.7744 \\
        & T2W & 0.0902 & 0.1006 & 21.24 & 0.7270 \\
        & FLAIR & 0.0856 & 0.1044 & 21.71 & 0.7448 \\

        \bottomrule
    \end{tabular}
\end{table}

\begin{figure}[!t]
	\centering
	\includegraphics[width=1\textwidth]{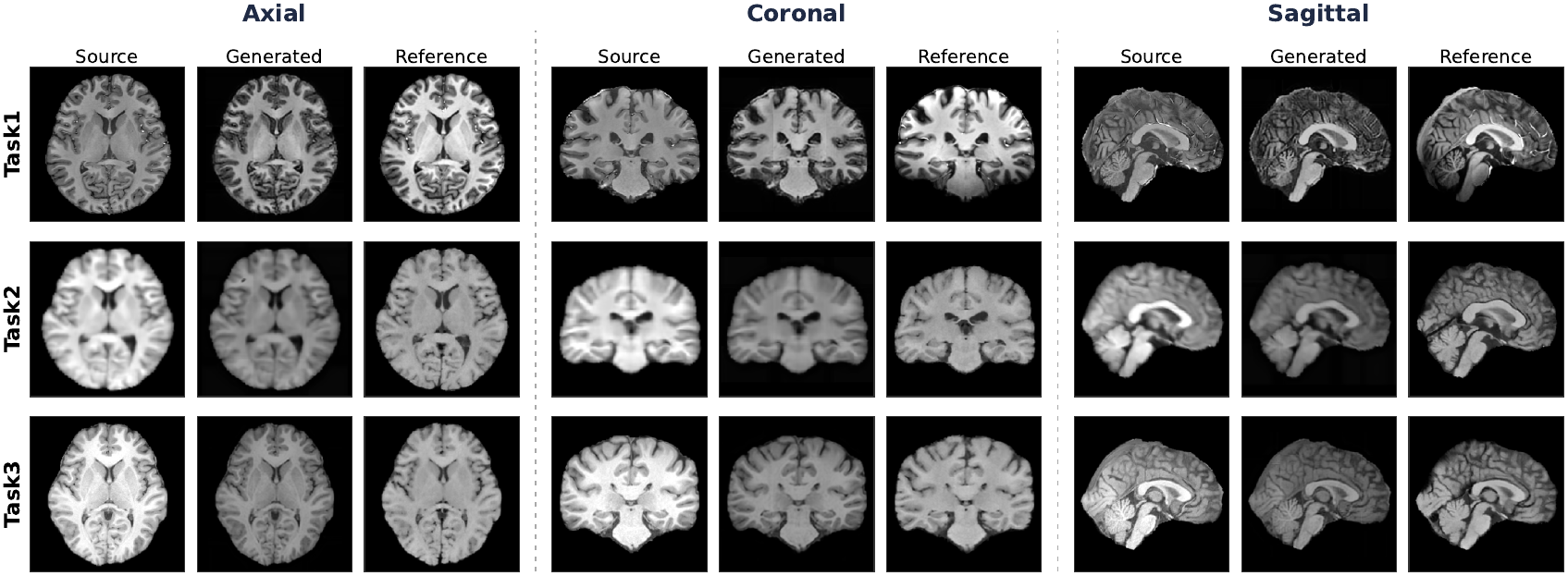}
	\caption{T1W qualitative full-volume examples in three orthogonal views. Each row shows source, generated, and paired reference slices in axial, coronal, and sagittal views. Representative directions are 5T to 7T for Any-to-7T, 0.1T to 1.5T for 0.1T-to-High, and 3T to 1.5T for Any-to-Any.}
	\label{fig:qualitative}
\end{figure}

The results show consistent but task-dependent volume performance.
Among the three objectives, Any-to-Any gives the most favorable average trade-off.
Task 1 is strongest for T2W, whereas FLAIR remains the most challenging modality because fluid-suppressed high-field contrast is difficult to infer from unpaired data.
Task 2 shows a similar modality dependence: T1W achieves the highest structural score, but the 0.1T-to-7T direction remains difficult across modalities.
In Task 3, high-to-low conversion is more stable than low-to-high conversion, reflecting the difficulty of recovering high-field detail from low-field inputs.
For visual assessment, Fig.~\ref{fig:qualitative} shows selected T1W examples.
The three orthogonal views indicate that the translated volumes preserve gross anatomy, with most visible errors arising from residual contrast mismatch in difficult low-to-high translations.

\subsection{Ablation Study}
Table~\ref{tab:ablation} summarizes a T1W loss-ablation study on representative Any-to-Any translation directions.
Each ablated model is briefly continued from the same pretrained backbone and evaluated on the same four paired testing cases.
The full objective provides a stable reference across error and structure metrics.
Removing cycle reconstruction increases nRMSE and MAE and reduces SSIM, confirming the importance of bidirectional reconstruction for anatomical anchoring.
Removing identity regularization has a smaller mixed effect, suggesting partial redundancy with the other consistency losses.
Content/style consistency is most important for structural fidelity: without it, SSIM is lowest and MAE is highest despite competitive nRMSE and PSNR.
Diversity regularization has a modest effect, while the full objective remains balanced across the four metrics.

\begin{table}[!t]
    \centering
    \caption{Loss ablation study on a representative T1W Any-to-Any testing subset.}
    \begin{tabular}{lcccc}
        \toprule
        Variant & nRMSE $\downarrow$ & MAE $\downarrow$ & PSNR $\uparrow$ & SSIM $\uparrow$ \\
        \midrule
        Full objective & 0.0825 & 0.0865 & 22.06 & 0.7748 \\
        w/o cycle reconstruction & 0.0845 & 0.0902 & 21.87 & 0.7300 \\
        w/o identity regularization & 0.0828 & 0.0878 & 22.06 & 0.7576 \\
        w/o content/style consistency & 0.0880 & 0.0913 & 21.45 & 0.7437 \\
        w/o diversity regularization & 0.0851 & 0.0885 & 21.86 & 0.7616 \\
        \bottomrule
    \end{tabular}
    \label{tab:ablation}
\end{table}

\section{Conclusion}\label{sec:conclusion}
We presented a task-adaptive 3D cross-field MRI translation framework based on field-conditioned content-style pretraining.
By learning Any-to-Any synthesis before task-specific adaptation, the method reuses a shared anatomical and contrast representation.
This representation supports Any-to-7T, 0.1T-to-High, and Any-to-Any.
Paired testing shows that the unified field-to-field setting provides the most consistent average performance among the evaluated objectives.
Task-specific fine-tuning further targets the MRIxFields challenge endpoints.
The current framework remains limited by unpaired low-to-high synthesis.
This limitation is most visible for 0.1T-to-7T and FLAIR translation, where residual contrast mismatch may occur.
Future work will investigate anatomical priors, overlap-aware inference, and larger-scale ablations.
These extensions may improve high-field detail recovery while preserving subject-specific structure.

\begin{credits}

\subsubsection{\discintname}
The authors have no competing interests to declare that are relevant to the content of this article.
\end{credits}

\bibliographystyle{splncs04}
\bibliography{Paper}

\end{document}